%% file: main.tex
\documentclass[runningheads]{llncs}
\usepackage[year=2024]{eccv}
\usepackage{multirow}
\usepackage{array}
\usepackage{pifont}
\usepackage{xspace}
\usepackage{tabularx}
\usepackage{eccvabbrv}
\usepackage{graphicx}
\usepackage{booktabs}
\usepackage{amsfonts}
\usepackage{amsmath}
\usepackage{booktabs}
\usepackage{float}
\usepackage{graphicx}
\usepackage{mathtools}
\usepackage{microtype}
\usepackage{multirow}
\usepackage{nicefrac}
\usepackage{pifont}
\usepackage{url}
\usepackage{wrapfig}
\usepackage{xcolor}
\usepackage{xspace}
\usepackage{tikz-cd}
\usepackage[accsupp]{axessibility}
\usepackage[colorlinks]{hyperref}
\usepackage{orcidlink}

\newcommand{\cmark}{\ding{51}}%
\newcommand{\xmark}{\ding{55}}%

\newcommand{\method}{CoTracker\xspace}
\newcommand{\davg}{\ensuremath{\delta_\text{avg}}\xspace}
\newcommand{\davgocc}{\ensuremath{\delta_\text{avg}^\text{occ}}\xspace}
\newcommand{\davgvis}{\ensuremath{\delta_\text{avg}^\text{vis}}\xspace}

\newcommand{\ua}{$\uparrow$}

\makeatletter
\DeclareRobustCommand\onedot{\futurelet\@let@token\@onedot}
\def\@onedot{\ifx\@let@token.\else.\null\fi\xspace}
\def\eg{\emph{e.g}\onedot} 
\def\ie{\emph{i.e}\onedot}

\makeatother

\makeatletter
\renewcommand{\paragraph}{%
  \@startsection{paragraph}{4}%
  {\z@}{-0.5em}{-0.5em}%
  {\normalfont\normalsize\it}%
}
\makeatother

\title{\method: It is Better to Track Together}
\titlerunning{\method}
\author{Nikita Karaev\inst{1,2} \and
Ignacio Rocco\inst{1} \and
Benjamin Graham\inst{1} \and
Natalia Neverova\inst{1} \and
Andrea Vedaldi\inst{1} \and
Christian Rupprecht\inst{2}}
\authorrunning{N.~Karaev et al.}
\institute{Meta AI \and
Visual Geometry Group, University of Oxford \\
\url{https://co-tracker.github.io/} \\
\email{nikita@robots.ox.ac.uk}}

\begin{document}
\maketitle
\begin{center}
\centering
\captionsetup{type=figure}
\includegraphics[width=\textwidth]{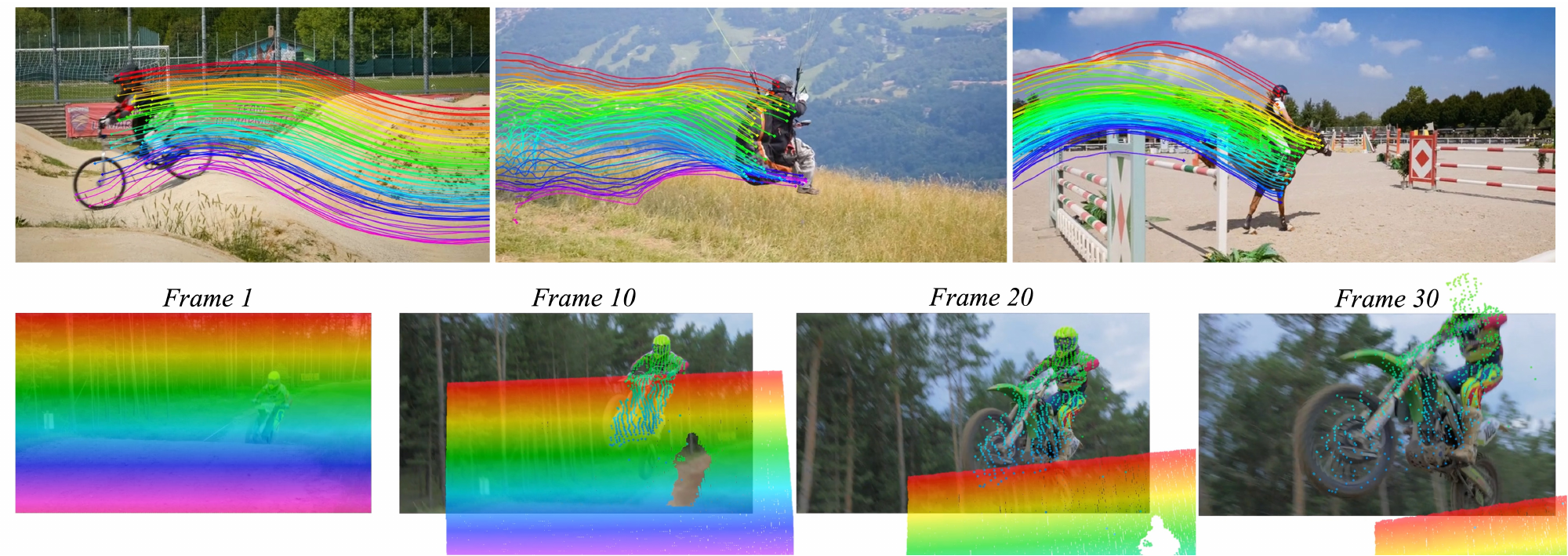}
\caption{\textbf{\method} is a new quasi-dense point tracker.
It can track 70k points jointly on a single GPU, exploiting dependencies between different tracks to enhance performance.
It has exceptional long-term tracking performance even in the presence of occlusions or when points leave the camera view (examples from DAVIS~\cite{pont20172017}, top row displays object tracks, bottom row shows displacements on a dense regular grid.
In frame 30, \method successfully tracks the driver's head even when it exists the camera view).}%
\label{fig:teaser}
\end{center}

\input{new/0_abstract}
\input{new/1_intro}
\input{new/2_related}

\input{new/3_method}
\input{new/4_experiments}

\input{new/5_limitations}
\input{new/6_conclusion}

\input{new/7_acknowledgments}

\bibliographystyle{splncs04}
\bibliography{vedaldi_general,vedaldi_specific,refs}
\input{supp_content}
\end{document}

%% file: new/0_abstract.tex
\begin{abstract}
We introduce \method, a transformer-based model that tracks a large number of 2D points in long video sequences.
Differently from most existing approaches that track points independently, \method tracks them jointly, accounting for their dependencies.
We show that joint tracking significantly improves tracking accuracy and robustness, and allows \method to track occluded points and points outside of the camera view.
We also introduce several innovations for this class of trackers, including using token proxies that significantly improve memory efficiency and allow \method to track 70k points jointly and simultaneously at inference on a single GPU\@.
\method is an online algorithm that operates causally on short windows.
However, it is trained utilizing unrolled windows as a recurrent network,  maintaining tracks for long periods of time even when points are occluded or leave the field of view.
Quantitatively, \method substantially outperforms prior trackers on standard point-tracking benchmarks.
 Code and model weights are available at \href{https://co-tracker.github.io/}{https://co-tracker.github.io/}
\keywords{Point tracking \and Motion estimation \and Optical flow}
\end{abstract}

%% file: new/1_intro.tex
\section{Introduction}%
\label{sec:intro}

We are interested in estimating point correspondences in videos containing dynamic objects and a moving camera.
Establishing point correspondences is an important problem in computer vision with many applications~\cite{agarwal2011building, teed2021droid, jiang2018super, yang2022banmo}.
There are two variants of this problem.
In \emph{optical flow}, the goal is to estimate the velocities of all points within a video frame.
This estimation is performed jointly for all points, but the motion is only predicted at an infinitesimal distance.
In \emph{point tracking}, the goal is to estimate the motion of points over an extended period of time.
For efficiency, trackers usually focus on a sparse set of points and treat them as statistically independent.
This is the case even for recent techniques such as TAPIR~\cite{doersch2023tapir} and PIPs++~\cite{zheng2023pointodyssey}, which employ modern architectures such as transformers and can track points even in the presence of occlusions.
However, points often have strong statistical dependencies (\eg, because they belong to the same object), which offers an opportunity for improvement.

We hypothesise that accounting for the dependency between tracked points can significantly improve tracking performance.
To explore this idea, we introduce \method, a new tracker that supports joint estimation of a very large number of tracks simultaneously, utilizing a transformer-based architecture and attention between tracks.
As we show in ablations, tracking points jointly significantly improves the tracking accuracy, especially when points are occluded.
In fact, we find that we can improve tracking quality further by adding \emph{more} points to the tracker than required by the user.
These additional \emph{support points} expand the context of the tracker, porting to point tracking the idea of using context, common in visual object tracking~\cite{jia2012visual, li2015reliable, song2017crest}.

In addition to joint tracking, \method incorporates additional architectural design innovations.
The network is a transformer operating in a sliding window fashion on a two-dimensional token representation, where dimensions are time and the set of tracked points~\cite{gberta_2021_ICML}.
Via suitable self-attention operators, the transformer considers each track as a whole for the duration of a window and can exchange information between tracks, thus exploiting their dependencies.
However, attention can be expensive when there are many tracks.
Hence, we introduce the concept of proxy tokens to point tracking that act similarly to registers~\cite{darcet2023vision, jaegle22perceiver, nagrani2021attention} while also reducing the memory complexity.
These tokens are processed as if they were a small number of additional tracks and allow the switch from expensive self-attention between tracks to efficient cross-attention between tracks and proxies.
In this way, \method can jointly track a \emph{near-dense set of tracks} on a single GPU\@ at inference.

\method is designed as an online tracker that operates on relatively short windows of frames.
Within a window, the tracks are initialized with queried points, and the network is tasked with progressively refining these initial estimations through iterative applications of the transformer.
The windows partially overlap and communicate, similar to a recurrent network.
Each subsequent overlapping window starts with refined predictions of the previous window and updates tracks for the new frames.
We optimize the recurrent application of the network via unrolled training, porting this concept from recurrent networks and visual object tracking~\cite{held2016learning, girshick2015deformable} to point tracking.
In this way,
\method achieves \emph{excellent long-term tracking performance} as well; in particular, joint and recurrent tracking allows points to be tracked through long occlusions.

We train \method on the synthetic TAP-Vid-Kubric~\cite{tapvid} and evaluate it on  TAP-Vid-\{DAVIS, RGB-Stacking\}, PointOdyssey~\cite{zheng2023pointodyssey} and DynamicReplica~\cite{karaev2023dynamicstereo}.
Our architecture works well for tracking single points and excels for groups of points, obtaining state-of-the-art tracking performance in several benchmarks compared to prior trackers.
In summary, our contributions are the following:
(i) we introduce the concept of \emph{joint point tracking} by sharing information between tracked points through an attention mechanism;
(ii) we propose to use \emph{support points} to provide additional context for point tracking;
(iii) we \emph{reduce the model's memory complexity} with proxy tokens; and
(iv) we propose an \emph{unrolled model training strategy}, which further improves tracking and occlusion accuracy.

\input{plot_single_joint}

%% file: plot_single_joint.tex
\begin{figure}[t]
\includegraphics[width=\textwidth]{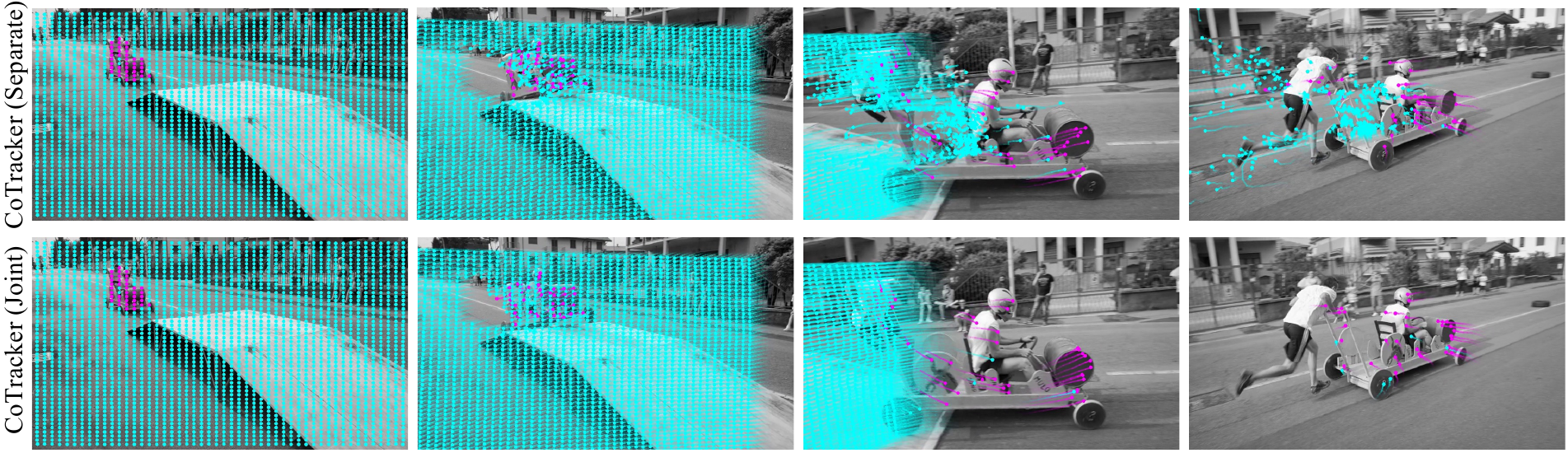}
\vspace{-2em}
\caption{\textbf{Non-joint tracking (top row) vs.~joint tracking (bottom).}  Background points are colored cyan while foreground points are magenta.
Non-joint point tracking causes background points to follow object  motion (cyan points in later frames), while tracking points together results in better tracks for foreground and background points.}%
\label{fig:single_joint}
\end{figure}

%% file: new/2_related.tex
\section{Related work}%
\label{sec:related}

\paragraph{Optical flow.}

Optical flow approximates dense instantaneous motion.
Originally approached by studying the brightness constancy equation~\cite{horn1981determining, lucas1981iterative, Black:ICCV:1993, bruhn2005lucas}, starting with
FlowNet~\cite{dosovitskiy2015flownet,ilg2017flownet} optical flow has been tackled using deep learning.
More recently, DCFlow~\cite{Xu_2017_CVPR} has introduced the computation of a 4D cost volume, used in most follow-up works~\cite{sun2018pwc, wang2020displacement, raft}.
Notable is RAFT~\cite{raft}, which introduced incremental flow updates and inspired several follow-up works~\cite{jiang2021learning, xu2021high, jiang2021learning, zhang2021separable}.
\method also uses 4D cost volumes and iterated updates but is applied to tracking.

Transformers~\cite{vaswani2017attention} have also been applied to the optical flow problem~\cite{huang2022flowformer, shi2023flowformer++, zhao2022global}.
Flowformer~\cite{huang2022flowformer} drew inspiration from RAFT and proposed a transformer-based approach that tokenizes the 4D cost volume.
GMFlow~\cite{zhao2022global} replaced the update network with a softmax with self-attention for refinement.
Perceiver IO~\cite{jaegle22perceiver} proposed a unified transformer for several tasks, including optical flow.

Optical flow can be applied to tracking by integrating predictions over time, but this approach accumulates errors in the form of drift~\cite{pips}, which motivates developing architectures like \method that track points over longer horizons.

\paragraph{Multi-frame optical flow.}

Several authors~\cite{ren2019fusion, shi2023videoflow, janai2018unsupervised, luo2023continuous} have extended optical flow to multiple frames.
Early methods used Kalman filtering~\cite{chin1994probabilistic, elad1998recursive} to encourage temporal consistency.
Modern multi-frame methods produce dense flow.
RAFT~\cite{raft} can be applied in a warm-start manner~\cite{sui2022craft, sun2022skflow} for multi-frame optical flow estimation.
However, these methods are not designed for long-term tracking and do not consider points occluded for a long time.
VideoFlow~\cite{shi2023videoflow} extends optical flow to three and five consecutive frames by integrating forward and backward motion features during flow refinement.
MFT~\cite{neoral2023mft} conducts optical flow estimation between distant frames and chooses the most reliable chain of optical flows.
Our tracker scales to produce semi-dense tracks.

\paragraph{Visual object tracking.}

Before the emergence of deep learning, some authors proposed handcrafted joint trackers~\cite{birchfield08joint,sand2008particle}, but few have considered doing so using deep networks like \method.
Our work is weakly related to \emph{multiple object tracking}~\cite{chen22visual} where tracking through occlusions~\cite{zhao2004tracking}, with changes in appearance~\cite{matthews2004template}, and with temporal priors~\cite{sidenbladh2000stochastic} have been extensively studied.

Some recent point trackers have been inspired by visual object tracking that predict the target states by extracting and combining features extracted from source and target image regions~\cite{bertinetto2016fully, danelljan2017eco, li2018learning, Bhat_2019_ICCV, danelljan2019atom}.
Transformers have also been applied in visual object tracking~\cite{Chen_2021_CVPR, cui2022mixformer}.
However, our focus is the tracking of points, including background points, not objects.

\paragraph{Tracking any point.}

Particle Video~\cite{sand2008particle} introduced the problem of Tracking Any Point in a video (TAP), and PIPs~\cite{pips} proposed a better model which can track through occlusions more reliably by predicting tracks in a sliding window and restarting them from the last frame where the point was visible.
However, PIP cannot track a point beyond the duration of a window.
TAP-Vid~\cite{tapvid} introduced a new benchmark and a simple baseline for TAP\@.
While the original Particle Video does track points jointly, PIPs and TAP-Vid track points independently (in parallel).
Inspired by TAP-Vid~\cite{tapvid} and PIPs~\cite{pips}, TAPIR~\cite{doersch2023tapir} further improved tracking with a two-stage, feed-forward point tracker that uses TAP-Vid-like matching and PIPs-like refinement.

PointOdyssey~\cite{zheng2023pointodyssey} addressed long-term tracking with PIPs++, a simplified version of PIPs, and introduced another benchmark for long-term tracking.
However, PIPs++ still tracks points independently.
OmniMotion~\cite{wang2023tracking} optimizes a volumetric representation for each video, refining correspondences in a canonical space.
However, this approach requires test-time optimization, which, due to its cost, is not suitable for many practical applications, especially online ones.

Trackers and optical flow models are often trained using synthetic datasets~\cite{karaev2023dynamicstereo, mayer2016large, dosovitskiy2015flownet, sun2021autoflow, butler2012naturalistic, zheng2023pointodyssey}, as annotating real data can be challenging.
Synthetic datasets provide accurate annotations, and training on them has demonstrated the ability to generalize to real-world data~\cite{mayer2016large, sun2021autoflow}.

%% file: new/3_method.tex
\section{\method}%
\label{sec:method}

Our goal is to track 2D points throughout the duration of a video $V = (I_t)_{t=1}^T$, which is a sequence of $T$ RGB frames $I_t \in \mathbb{R}^{3 \times H \times W}$.
The goal of the \emph{tracker} is to predict $N$ point tracks
$
P^i_t = (x^i_t, y^i_t) \in \mathbb{R}^{2}
$,
$
t =t^i,\dots,T
$,
$
i=1,\dots,N
$,
where
$
t^i \in \{1, \dots, T\}
$
is the time when the track starts.
The tracker also predicts the \emph{visibility flag} $v^i_t \in \{0,1\}$, which tells if a point is visible or occluded in a given frame.
To make the task definition unambiguous~\cite{tapvid}, we assume that each point is visible at the start of the track (\ie, $v^i_{t_i} = 1$).
The tracker is thus given as input the video $V$ and the starting locations and times $(P^i_{t^i}, t^i)_{i=1}^N$ of $N$ tracks, and outputs an estimate
$
(\hat P^i_t = (\hat x^i_t, \hat y^i_t), \hat v^i_t)
$
of the track locations and visibility for all valid times, \ie, $t \ge t^i$.

\input{sliding_window_inference_vis}
\subsection{Transformer formulation}

We implement the tracker as a transformer neural network (\cref{fig:window_architecture,fig:architecture})
$
\Psi : G \to O
$.
The goal of this transformer is to improve an initial estimate of the tracks.
Tracks are encoded as a grid $G$ of input tokens $G^i_t$, one for each track $i=1,\dots,N$, and time $t=1,\dots, T$.
The updated tracks are expressed by a corresponding grid $O$ of output tokens $O^i_t$.

\paragraph{Image features.}

We extract dense $d$-dimensional appearance features
$
\phi(I_t) \in \mathbb{R}^{d\times \frac{H}{k} \times \frac{W}{k}}
$
from each video frame $I_t$ using a convolutional neural network (trained end-to-end).
We reduce the resolution by $k=4$ for efficiency.
We also consider several scaled versions
$
\phi_s(I_t)\in \mathbb{R}^{d\times \frac{H}{k2^{s-1}} \times \frac{W}{k2^{s-1}}}
$,
of the features with strides $s=1,\dots,S$ and use $S=4$ scales.
These downscaled features are obtained by applying average pooling to the base features.

\paragraph{Track features.}

The appearance of the tracks is captured by feature vectors $Q_t^i \in \mathbb{R}^{d}$ (these are time-dependent to accommodate changes in the track appearance).
They are initialized by broadcasting image features sampled at the starting locations and are updated by the neural network, as explained below.

\paragraph{Spatial correlation features.}

In order to facilitate matching tracks to images, we adopt correlation features $C^i_t \in \mathbb{R}^S$ similar to RAFT~\cite{raft}.
Each $C^i_t$ is obtained by comparing the track features $Q_t^i$ to the image features $\phi_s(I_t)$ around the current estimate $\hat P_t^i$ of the track's location.
Specifically, the vector $C_{t}^i$ is obtained by stacking the inner products
$
[C_t^i]_{s\delta}
=
\langle
Q^i_t,~
\phi_s(I_t)[
  \hat P_t^i / ks + \delta
]
\rangle,
$
where
$
s = 1,\dots, S
$
are the feature scales
and
$
\delta \in \mathbb{Z}^2
$,
$
\| \delta \|_\infty \leq \Delta
$
are offsets.
The image features $\phi_s(I_t)$ are sampled at non-integer locations by using bilinear interpolation and border padding.
The dimension of $C^{i}_t$ is $\big (2\Delta+1)^2 S=196$ for our choice $S = 4; \Delta = 3$.

\paragraph{Tokens.}

The input tokens $G(\hat P, \hat v, Q)$ code for position, visibility, appearance, and correlation of the tracks.
They are given by the following concatenation of features with added positional encodings:
\begin{equation}
   G^i_t = (
   \hat P^i_t - \hat P^i_1,~
   \hat v^i_t,~
   Q^i_t,~
   C^i_t,~
   \eta (\hat P^i_t - \hat P^i_1)
   )
   + \eta'(\hat P^i_1) + \eta'(t).
\end{equation}
All the components except the last one have been introduced above.
The last component is derived from the estimated position: it is the sinusoidal positional encoding $\eta$ of the track location with respect to the initial location at time $t=1$.
We also add encodings $\eta'$ of the start location $P^i_1$ and for the time $t$, with appropriate dimensions.
In fact, we found it beneficial to separately encode the position of points at the first frame and their relative displacement to this frame.

The output tokens $O(\Delta \hat P, \Delta Q)$ contain updates for location and appearance, \ie
$
O^i_t =
(
    \Delta \hat P^{i}_t,
    \Delta Q^{i}_t
).
$

\paragraph{Iterated transformer application.}%
\label{sec:updates}

We apply the transformer $M$ times in order to progressively improve the track estimates.
Let $m=0,1,\dots,M$ index the estimate, with $m=0$ denoting initialization.
Each update computes
$$
  O(\Delta \hat P, \Delta Q)
  = \Psi(G(
    \hat P^{(m)}, \hat v^{(0)}, Q^{(m)}
  ))
$$
and sets
$
\hat P^{(m+1)} = \hat P^{(m)} + \Delta \hat P
$
and
$
Q^{(m+1)} = Q^{(m)} + \Delta Q.
$
The visibility mask $\hat v$ is not updated iteratively, but only once after the last transformer application as
$
 \hat v^{(M)} = \sigma(W Q^{(M)})
$,
where $\sigma$ is the sigmoid activation function and $W$ is a learned matrix of weights.
We found that updating the visibility flag iteratively did not improve performance, likely due to the fact that predicting visibility requires predicting an accurate location first.

For $m=0$, the position, visibility and appearance estimates $\hat P^{(0)}$, $v^{(0)}$ and $Q^{(0)}$ are initialized by broadcasting their query point's initial location $P^i_{t^i}$, visibility $v^i_{t^i}=1$ (meaning visible) and appearance $\phi(I_{t^i})[P^i_{t^i}/k]$ to all times $t=1,\dots,T$.

\input{architecture}

\subsection{Transformer architecture and proxy tokens}%
\label{s:virtual_tracks}
The transformer $\Psi$ interleaves attention layers operating across the \emph{time} and \emph{track dimensions}, respectively.
Factorising the attention~\cite{gberta_2021_ICML} across time and tracks makes the model computationally tractable:
the complexity is reduced from $O(N^2T^2)$ to $O(N^2+T^2)$.
However, for very large values of $N$, this cost is still prohibitive.
We thus propose a new design, in which we introduce $K$ \emph{proxy tracks}, where $K \ll N$ is a hyper-parameter.
% Proxy tracks are formed by concatenating rows of fixed learnable tokens to the list of actual tracks at the input to the transformer.
Proxy tracks are learned fixed tokens that we concatenate to the list of `regular' tracks at the input to the transformer and discard at the output.
% , and removing them at the output.

% We concatenate the learned proxy tokens to the list of actual tracks at the input to the transformer.

For time attention, `regular' and proxy tracks are processed identically.
For track attention, however, regular tracks cross-attend the proxies, but not each other, reducing the cost to $O(NK + K^2 + T^2)$.
 We discard the proxy tracks at the output of the transformer. 

Proxies are often used to accelerate, \eg, large graph neural networks; here, proxies are learnable query tokens.
In computer vision, DETR~\cite{carion20end-to-end} makes use of learnable queries in a transformer and methods like {}\cite{jaegle22perceiver} use them to decode dense outputs.
Our proxy tokens are different: They mirror and shadow the computation of the regular tracks, similarly to registers~\cite{darcet2023vision}.

\subsection{Windowed inference and unrolled training}%
\label{sec:unrolled_training}

An advantage of formulating tracking as the progressive refinement of a window of tracks is the ability to process arbitrarily long videos:
It suffices to initialize the next window using partially overlapping tracks from the previous one.

%  of an initial track estimate is  updates allows to process arbitrarily long videos sliding window application to process ,
% by initializing the iterative transformer updates with the outputs of the previous time window.

Consider, in particular, a video $V$ of length $T' > T$ longer than the maximum window length $T$ supported by the architecture.
To track points throughout the entire video $V$, we split the video in
$J = \left\lceil 2T'/T - 1 \right\rceil$ windows of length $T$, with an overlap of $T/2$ frames.\footnote{We assume that $T$ is even. The last window is shorter if $T/2$ does not divide $T'$.}

Let the superscript $(m,j)$ denote the $m$-th application of the transformer to the $j$-th window.
We thus have a $M \times J$ grid of quantities
$
(
  \hat P^{(m,j)},~
  \hat v^{(m,j)},~
  Q^{(m,j)}
)
$,
spanning transformer iterations and windows.
For $m=0$ and $j=1$, these are initialized as for the single window case.
Then, the transformer is applied $M$ times to obtain the estimate $(M,1)$.
The latter is used to initialize estimate $(0,2)$ via broadcasting.
Specifically, the first $T/2$ components of $\hat P^{(0,2)}$ are copies of the last $T/2$ components of $\hat P^{(M,1)}$; the last $T/2$ components of $\hat P^{(0,2)}$ are instead copies of the last time $t=T/2-1$ from $\hat P^{(M,1)}$.
The same update rule is used for
$
\hat v^{(0,2)}
$,
while
$
Q^{(0,j)}
$
is always initialized with the features $Q$ of the track's staring location.
This process is repeated until estimate $(M,J)$ is obtained.

The windowed transformer essentially operates as a recurrent network, so we train it in an unrolled fashion.
Specifically, we optimize the track prediction error summed over iterated transformer applications and windows
\begin{equation}\label{e:loss1}
\mathcal{L}_1(\hat{P}, P)
=
\sum_{j=1}^{J}
\sum_{m=1}^{M}
\gamma^{M-m}
\|
\hat P^{(m,j)} - P^{(j)}
\|,
\end{equation}
where $\gamma=0.8$ discounts early transformer updates.
Here, $P^{(j)}$ contains the ground-truth trajectories restricted to window $j$ (trajectories which start in the middle of the window are padded backwards).
The second loss is the cross entropy of the visibility flags
$
\mathcal{L}_2(\hat{v}, v)
=
\sum_{j=1}^{J}
\operatorname{CE}(
  \hat v^{(M,j)},
  v^{(j)}
)
$.
While only a moderate number of windows are used in the loss during training due to the computational cost, at test time, we can unroll the windowed transformer applications arbitrarily, thus, in principle, handling any video length. In combination with joint tracking, unrolling allows the model to track points through occlusions of duration longer than one sliding window and to track points even outside of the camera view.

\subsection{Support points}%
\label{sec:point_selection}

\method can take advantage of tracking several points jointly.
It is typical for applications to have several points to track, but in some cases, one might be interested in tracking a few or even a single point at inference time.

In these cases, we found it beneficial to track additional \textit{support points} which are not explicitly requested by the user.
Moreover, we have found that different configurations of such support points can lead to small differences in performance.
We experiment with various configuration types, visualized in \cref{fig:evaluation_sampling_fig}.
With the ``global'' strategy, the support points form a regular grid across the whole image.
With the ``local'' strategy, the grid of points is centred around the point we wish to track, thus allowing the model to focus on a neighbourhood of it.
Note that these patterns are only considered at inference time and are used to improve the tracker's accuracy for the target points by incorporating context.

%% file: sliding_window_inference_vis.tex
\begin{figure*}[tb]
  \centering
  \includegraphics[width=\textwidth]{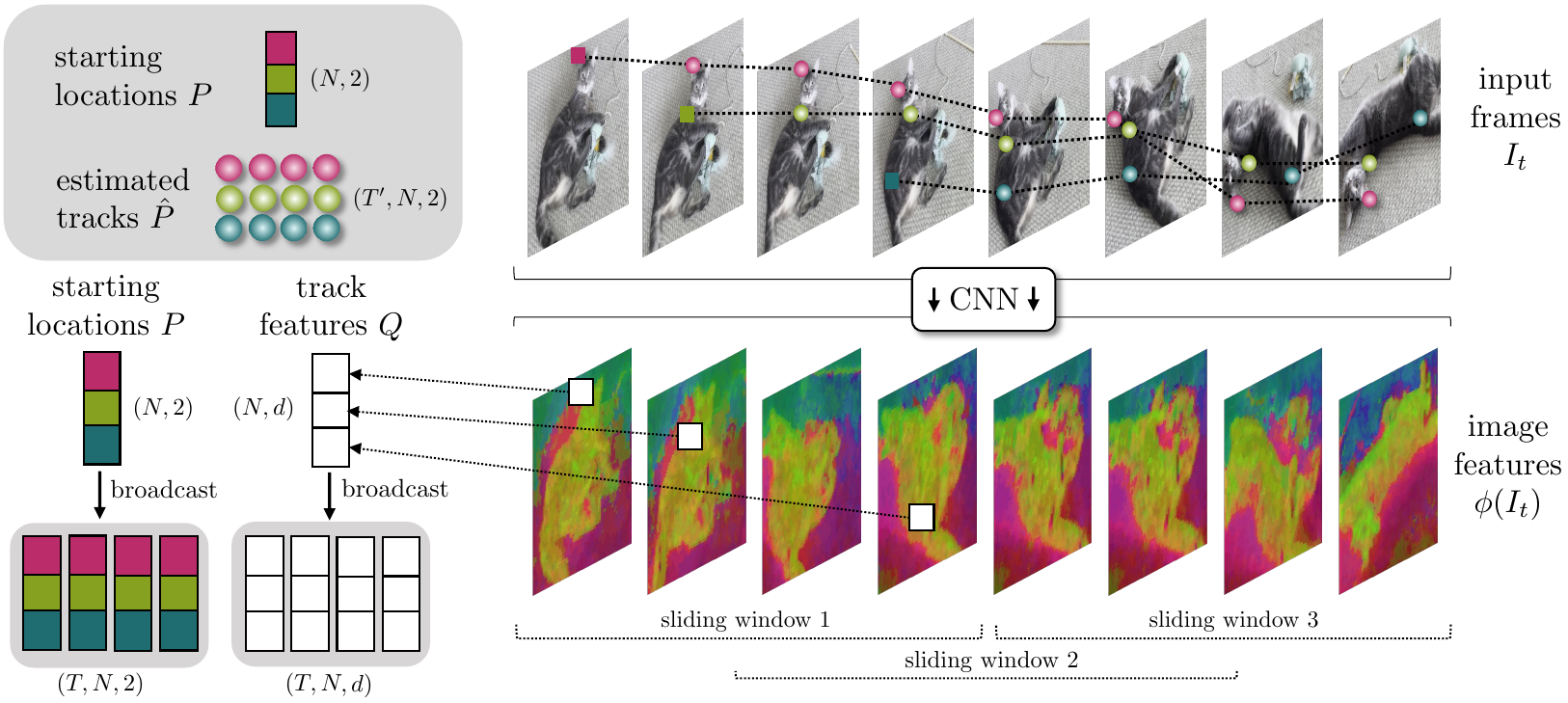}
  % \fbox{\rule[-.5cm]{0cm}{4cm} \rule[-.5cm]{4cm}{0cm}}
  \caption{\textbf{CoTracker architecture.} We compute convolutional features $\phi(I_t)$ for every frame and process them with sliding windows. To initialize track features $Q$, we bilinearly sample from $\phi(I_t)$ with starting point locations $P$. Locations $P$ also serve to initialize estimated tracks $\hat P$. See \cref{fig:architecture} for a visualization of one sliding window.}\label{fig:window_architecture}
\end{figure*}

%% file: architecture.tex
\begin{figure*}
  \centering
  \includegraphics[width=\textwidth]{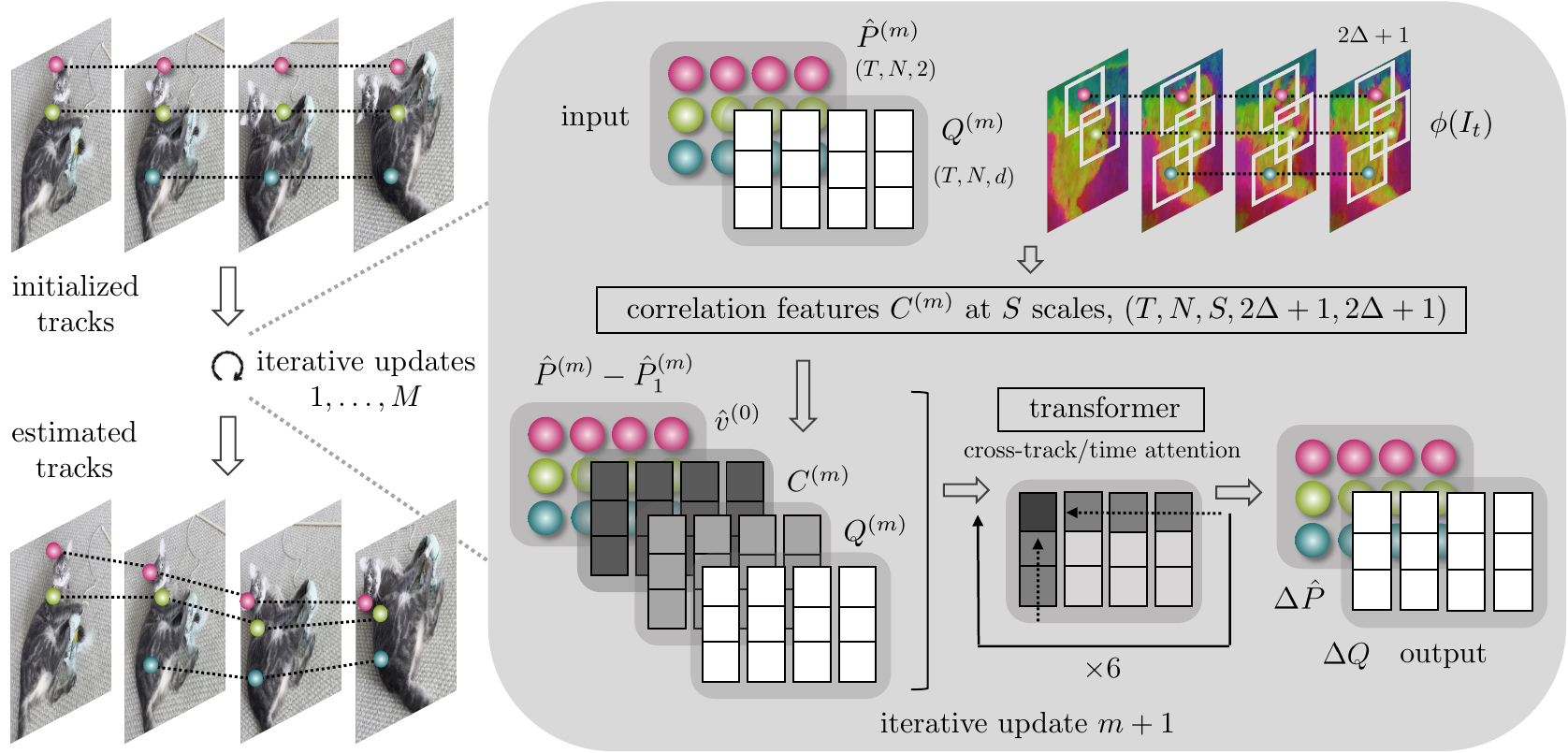}
  
  \caption{\textbf{CoTracker architecture.}  Visualization of one sliding window with $M$ updates. During one iteration, we update point tracks $\hat P^{(m)}$ and track features $Q^{(m)}$. $Q^{(0)}$ is initialized with the initially sampled features $Q$ for all sliding windows, $\hat P^{(0)}$ with the starting locations for the first window. For other windows, $\hat P^{(0)}$ starts with predictions for frames processed in the preceding sliding window, and with the last predicted positions for the unseen frames.  
  We compute visibility $\hat v$ after the last update $M$.
  }\label{fig:architecture}
\end{figure*}

%% file: new/4_experiments.tex
\section{Experiments}%
\label{sec:experiments}

\input{evaluation_sampling}
\input{cotracker_pips_vis}

We evaluate \method on several standard real and synthetic tracking benchmarks.
After discussing implementation details, datasets and benchmarks, we compare \method to the state of the art (\cref{sec:sota}) and ablate our design, demonstrating the importance of joint tracking and other innovations (\cref{sec:ablations})

\paragraph{Datasets and benchmarks.}

We train the model on \textbf{TAP-Vid-Kubric}~\cite{tapvid}, as in~\cite{tapvid,doersch2023tapir}.
It consists of sequences of 24 frames showing 3D rigid objects falling to the ground under gravity and bouncing, generated using the Kubric engine~\cite{greff2022kubric}.
Point tracks are selected randomly, primarily on objects and some on the background.
Objects occlude each other and the background as they move.

For evaluation, we consider \textbf{TAP-Vid-DAVIS}~\cite{tapvid}, containing 30 real sequences of about 100 frames, and \textbf{TAP-Vid-RGB-Stacking}~\cite{tapvid} with 50 synthetic sequences with objects being moved by robots of 200--300 frames each.
Points are queried on random objects at random times and evaluation assesses both predictions of positions and visibility flags.
This uses two evaluation protocols.
In the ``first'' protocol, each point is queried only once in the video, at the first frame where it becomes visible.
The tracker is expected to operate causally, predicting positions only for future frames.
In the ``strided'' protocol, points are queried every five frames, and tracking is bidirectional.
Given that most trackers (ours, PIPs, PIPs++) are causal, we run them twice, on the video and its reverse.
We assess tracking accuracy using the TAP-Vid metrics~\cite{tapvid}:
Occlusion Accuracy (OA\@; accuracy of occlusion prediction treated as binary classification),
\davgvis (fraction of visible points tracked within 1, 2, 4, 8 and 16 pixels, averaged over thresholds), and
Average Jaccard (AJ, measuring jointly geometric and occlusion prediction accuracy).
Following~\cite{tapvid}, the \davgvis and AJ tracking thresholds are applied after virtually resizing images to $256\times 256$ pixels.
Note that TAP-Vid benchmarks evaluate only \textit{visible} points.

We also evaluate on \textbf{PointOdyssey}~\cite{zheng2023pointodyssey}, a more recent synthetic benchmark dataset for long-term tracking.
It contains 100 sequences that are several thousand frames long, with objects and characters moving around the scene.
Scenes are much more realistic than TAP-Vid-Kubric.
We train and evaluate \method on PointOdyssey and report \davgvis and \davgocc, \davg.
The last two are the same as \davgvis, but for occluded and all points, respectively;
they can be computed because the dataset is synthetic so the ground-truth positions of invisible points are known.
We also report their \emph{Survival} rate, which is the average fraction of video frames until the tracker fails (detected when the tracking error exceeds 50 pixels).

\textbf{Dynamic Replica}~\cite{karaev2023dynamicstereo} is a synthetic dataset for 3D reconstruction that contains long-term tracking annotations.
It consists of 500 300-frame sequences of articulated models of people and animals.
We evaluate models trained on Kubric and measure \davgvis and \davgocc on the ``valid'' split of this dataset.

\paragraph{Implementation details.}

Here we provide important implementation details and refer to the sup.~mat.~for others.
We \textbf{train} the model for 50,000 iterations on 6,000 TAP-Vid-Kubric sequences of $T'=24$ frames using sliding window size $T=8$, which takes 40 hours.
We use $32$ NVIDIA A100 80GB GPUs with a batch size of 1.
Training tracks are sampled preferentially on objects.
During training, we construct batches of $N=768$ tracks with visible points either in the first or middle frames of the sequence to train the tracker to handle both cases.
In a similar manner, we also train a second version of \method on the train split of PointOdyssey, using 128 tracks randomly sampled from sequences of length $T'=56$.

A technical difficulty is that the number of tracks $N$ varies from window to window, as new points can be added to the tracker at any time.
While conceptually, this is handled by simply adding more tokens when needed, in practice, changing the number of tokens makes batching data for training difficult.
Instead, we allocate enough tokens for all the tracked points, regardless of which window they are added in, and use masking to ignore tokens that are not yet used.

The video \textbf{resolution} can also affect tracking performance.
For evaluation on TAP-Vid, we follow their protocol and downsample videos to $256\times 256$, which ensures that no more information than expected is passed to the tracker.
However, each tracker has a different native resolution at which it is trained ($384 \times 512$ for PIPs and \method; $512 \times 896$ for PIPs++).
We thus resize videos from $256\times 256$ to the native resolution of each tracker before running it to ensure fairness.
For Dynamic Replica and PointOdyssey, we resize videos to the native tracker resolution, as there is no prescribed resolution.

\input{tabs/tab_comparison}
\input{tabs/tab_comparison_pointodyssey}

\subsection{Comparisons to the State of the Art}%
\label{sec:sota}

We compare \method to state-of-the-art trackers on TAP-Vid-DAVIS and TAP-Vid-RGB-Stacking in \cref{tab:tab_comparison}.
For fairness, we use the same configuration of support points identified in the ablations (see \cref{sec:ablations}) across all benchmarks.
We also evaluate \method on PointOdyssey in \cref{tab:tab_comparison_pointodyssey} and on Dynamic Replica in \cref{tab:tab_comparison_dr} by tracking jointly all the target points.
\method improves most metrics by substantial margins on the TAP-Vid benchmarks while generalizing well from synthetic Kubric to real DAVIS\@.
\method also improves tracking accuracy \davgvis in all cases.
On PointOdyssey, despite using a window size of only 8 frames, \method has a better survival rate than PIPs++, which uses a 128-frame sliding window.
This shows the power of unrolled training, which trains the model to propagate information across a long series of sliding windows.
The gap between \method and other methods in \cref{tab:tab_comparison_pointodyssey} and \cref{tab:tab_comparison_dr} for \davgocc is higher than for \davgvis, which shows that \method excels in tracking occluded points thanks to jointly tracking groups of points.

\subsection{Ablations}%
\label{sec:ablations}

\paragraph{Tracking together is better.}

In \cref{tab:tab_joint_tracking_ablation}, we demonstrate the importance of tracking points jointly, which is the main motivation of \method.
This is achieved by removing cross-track attention from the tracker, thus ignoring the dependencies between tracks entirely.
For fairness, rather than simply removing the six cross-track attention layers, we maintain the same model size and replace them with twelve time-attention layers, resulting in a comparable model size.
On Dynamic Replica, joint tracking improves the accuracy of occluded points \davgocc 28.8 $\rightarrow$ 37.6 (+30.6\%) more than the accuracy of visible points \davgvis 62.4 $\rightarrow$ 68.9 (+10.4\%), showing the effectiveness of joint tracking in understanding the scene motion, including currently invisible points.

\input{tabs/tab_space_time_ablation_new}

\paragraph{Importance of unrolled training.}

We design \method for tracking in a windowed manner, and in \cref{tab:tab_unrolling}, we thus assess the effect of unrolling windows during training (\cref{sec:unrolled_training}).
Switching off unrolled training decreases performance by 18 AJ points.
Hence, unrolled training helps to track over long periods of time, $>10\times$ longer than the sequences used in training.

\paragraph{Effect of proxy tokens on scalability.}

In \cref{tab:tab_ablate_clusters}, we assess the benefits of using proxy tokens during inference.
For a fixed memory budget (80 GB), using proxy tokens instead of full self-attention allows us to track $\times 7.4$ more points than without them;
in fact, we can track a whole 263 $\times$ 263 grid, which is quasi-dense for the input video resolution.
Moreover, proxy tokens also reduce the time complexity of the model and make it $\times 7$ faster at inference time for the maximum number of tracked points.
The number of proxy tokens does not affect scalability but affects performance, with the best results obtained using 64 proxy tokens.
In short, \emph{proxy tokens allow to track almost one order of magnitude more tracks with higher accuracy than the naive self-attention}.

\begin{table}[tb]
    \begin{minipage}{.50\linewidth}
        \centering
        \setlength\tabcolsep{0.9pt}
       \input{tabs/tab_ablate_clusters}

    \end{minipage}%
    \hspace{0.02\linewidth}
    \begin{minipage}{0.45\linewidth}
        \centering
        \setlength\tabcolsep{1pt}
        \input{tabs/tab_space_time_ablation}

    \end{minipage}
\end{table}

\paragraph{Optimal support point configurations.}

We compare the effect of choosing different configurations of support points from \cref{fig:evaluation_sampling_fig} in \cref{tab:tab_joint_tracking_ablation_global_local}.
In order to do so, we take a single benchmark point at a time and add additional support points (\cref{sec:point_selection}) to allow the model to perform joint tracking.
This protocol, where a single benchmark point is tracked each time, is also important for fairness when compared to the state-of-the-art.
This is because the benchmark points might be biased for objects, which is particularly evident in TAP-Vid-DAVIS (\cref{fig:evaluation_sampling_fig}).
Passing more than one such biased point at the same time to a joint tracker like ours could help it by revealing the outline of objects to the tracker.
By considering one benchmark point at a time, we ensure that no such ground-truth information can leak into it.
We use this scheme for the TAP-Vid benchmarks, which explicitly prohibits tracking more than one ground-truth point at a time.

Adding any configurations of support points helps, but the local configuration helps much more than the global, presumably because the model uses dependencies with other points located on the same object.
The combination of global and local context works best, likely because the model can track both the camera motion and the object motion in this case.

\input{figure_lab}

%% file: evaluation_sampling.tex
\begin{figure*}[t]
\centering
\includegraphics[width=\textwidth]{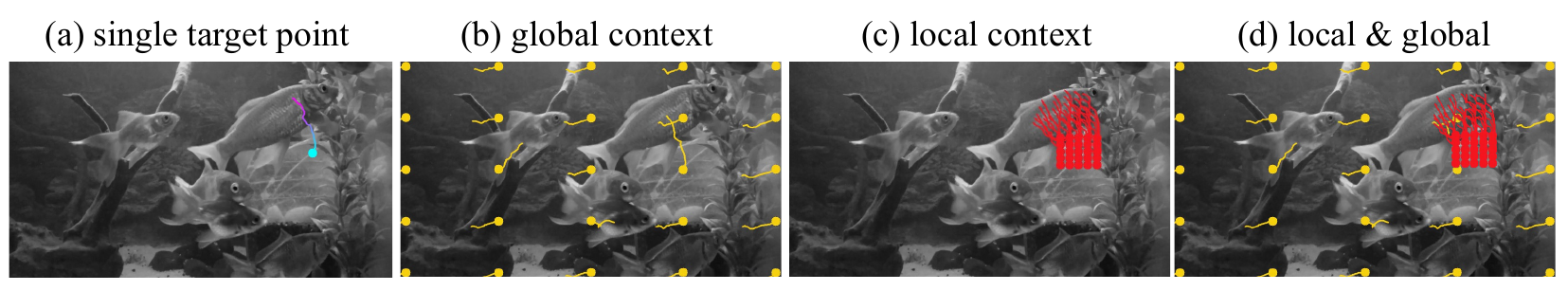}
\caption{\textbf{Tracked and support points.}
We visualize different configurations of points along with their tracks. Points represent the start of each track, which clearly illustrates grids of points in (b) and (c).
(a) A single target point in a sequence from TAP-Vid-DAVIS\@.
(b) A global grid of support points.
(c) A local  grid of support points.
(d) Local and global support points.
In \cref{tab:tab_joint_tracking_ablation_global_local} we evaluate CoTracker by combining different configurations of support points, as described in \cref{sec:point_selection}.
}%
\label{fig:evaluation_sampling_fig}
\end{figure*}

%% file: cotracker_pips_vis.tex
\begin{figure*}[t]
\centering
\includegraphics[width=\linewidth]{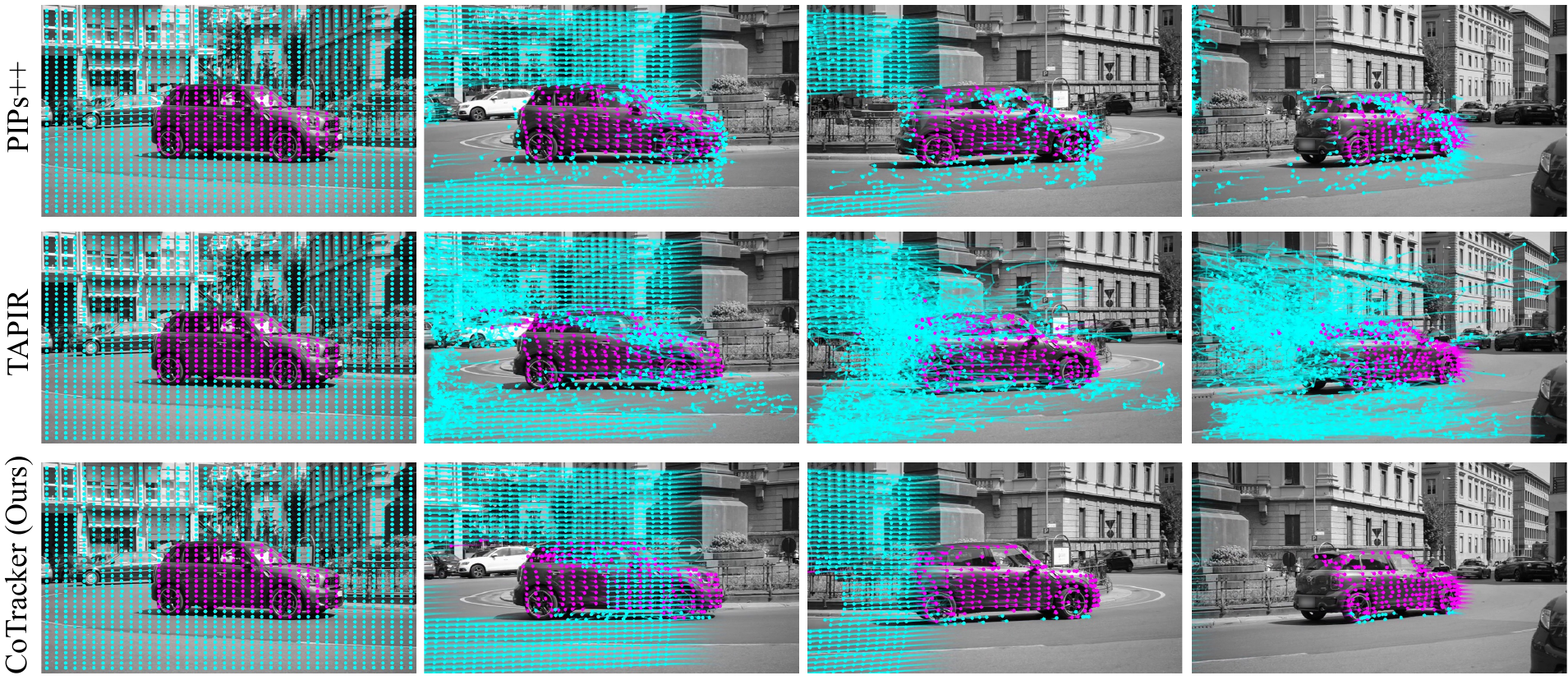}
% \fbox{\rule[-.5cm]{0cm}{4cm} \rule[-.5cm]{4cm}{0cm}}
\caption{\textbf{Qualitative Results}.
For PIPs++ (top), many points are incorrectly tracked and end up being `stuck' on the front of the car.  TAPIR (middle) works well for visible points but fails to handle occluded points. As soon as a point becomes occluded, it starts moving chaotically around the image.
Our \method (bottom) produces cleaner tracks. The tracks are also more `linear' than those of  PIPs++ or TAPIR, which is accurate as the primary motion is a homography (the observer does not translate).}
\end{figure*}

%% file: tabs/tab_comparison.tex
\begin{table*}[t]
\centering
\footnotesize
\begin{tabular}{llcccccccccccc}
\toprule
\multirow{2}{*}{Method} &\multirow{2}{*}{Train} & \multicolumn{3}{c}{DAVIS First} & \multicolumn{3}{c}{DAVIS Strided} & \multicolumn{3}{c}{RGB-S First}  \\
% & \multicolumn{3}{c}{RGB-S every 5th}
\cmidrule(lr){3-5}
\cmidrule(lr){6-8}
\cmidrule(lr){9-11}
 & & AJ \ua & \davgvis\ua & OA \ua & AJ \ua & \davgvis\ua & OA \ua &  AJ \ua & \davgvis\ua & OA\ua \\
 % &  AJ \ua & \davgvis\ua & OA\ua 
\midrule
TAP-Net~\cite{tapvid} & K & 33.0 & 48.6 & 78.8 & 38.4 & 53.1 & 82.3 & --- & --- & --- \\
OmniMotion~\cite{wang2023tracking} & --- & 52.8 & 66.9 & \underline{87.1} & 51.7 & 67.5 & 85.3 &  \underline{69.5} &  \underline{82.4} &  \textbf{90.3} \\
PIPs~\cite{pips} &  FT++ & 42.2 & 64.8 & 77.7 & 52.4 & 70.0 & 83.6&  ---  & 59.1 & ---  \\
MFT~\cite{neoral2023mft} & K++ & 47.3 & 66.8 & 77.8 & 56.1 & 70.8 & 86.9 & --- &  --- &  --- \\
PIPs++~\cite{zheng2023pointodyssey}  & PO & --- & 69.1 & --- & --- & \underline{73.7} & --- &  ---  & 77.8 & --- & \\
TAPIR~\cite{doersch2023tapir} & K & \underline{56.2} & \underline{70.0} & 86.5 & \underline{61.3} & 73.6 & \underline{88.8} &  60.3 & 74.1 & 85.5 \\
\midrule
CoTracker (Ours) & K & \textbf{62.2 }& \textbf{75.7} & \textbf{89.3} & \textbf{65.9} & \textbf{79.4} & \textbf{89.9} & 
 \textbf{71.6} & \textbf{83.3}  & \underline{89.6} \\
 
\bottomrule
\end{tabular}
\caption{
\textbf{TAP-Vid benchmarks}
We compare \method to the best trackers available on TAP-Vid benchmarks utilizing both their ``first'' and ``strided'' evaluation protocols. 
During evaluation, TAPIR and OmniMotion have access to all video frames at once, while we process videos in an online manner using a causal sliding window. This puts online trackers like ours at a disadvantage on long videos with slow motion, such as in RGB-S. To address this, we decrease the frame rate for RGB-S evaluation by keeping every 5th frame.
Training data:
(K) Kubric,
(K++) Kubric+more,
(FT) FlyingThings++,
(PO) Point Odyssey.
 }\label{tab:tab_comparison}
\end{table*}

%% file: tabs/tab_comparison_pointodyssey.tex
\begin{table}[t]
\begin{minipage}{.54\linewidth}
\centering
\setlength\tabcolsep{0.9pt}
\begin{tabular}{lcccc}
\toprule
\multirow{2}{*}{Method} & \multicolumn{4}{c}{PointOdyssey} \\
\cmidrule(lr){2-5}
                                              & \davg\ua      & \davgvis\ua   & \davgocc\ua   & Survival \ua  \\
\midrule
TAP-Net~\cite{tapvid}                         & 28.4          & ---           & ---           & 18.3          \\
PIPs~\cite{harley22particle}                  & 27.3          & ---           & ---           & 42.3          \\
PIPs++$^\dagger$~\cite{zheng2023pointodyssey} & 29.0          & 32.4          & 18.8          & 47.0          \\
\midrule
CoTracker (Ours)                              & \textbf{30.2} & \textbf{32.7} & \textbf{24.2} & \textbf{55.2} \\
\bottomrule
\end{tabular}
\caption{
\textbf{PointOdyssey}.
All methods are trained on PointOdyssey.
$^\dagger$ results for~\cite{zheng2023pointodyssey} obtained using released code and model.}%
\label{tab:tab_comparison_pointodyssey}
\end{minipage}%
~~~~%
\begin{minipage}{0.42\linewidth}
\centering
\setlength\tabcolsep{1pt}
\begin{tabular}{lccc}
\toprule
\multirow{2}{*}{Method} & \multicolumn{3}{c}{Dynamic Replica}\\
\cmidrule(lr){2-4}
                                    & \davg\ua      & \davgvis\ua   & \davgocc\ua   \\
\midrule
TAP-Net~\cite{tapvid}               & 45.5          & 53.3          & 20.0          \\
PIPs~\cite{harley22particle}        & 41.0          & 47.1          & 21.0          \\

PIPs++~\cite{zheng2023pointodyssey} & 55.5          & 64.0          & 28.5          \\
TAPIR~\cite{doersch2023tapir}       & 56.8          & 66.1          & 27.2          \\
\midrule
CoTracker (Ours)                    & \textbf{61.6} & \textbf{68.9} & \textbf{37.6} \\
\bottomrule
\end{tabular}
\caption{
\textbf{Dynamic Replica}.
The gap between \method and other methods in tracking occluded points' accuracy \davgocc is higher than \davgvis.}%
\label{tab:tab_comparison_dr}
\end{minipage}
\end{table}

%% file: tabs/tab_space_time_ablation_new.tex
\begin{table}[t]
\begin{minipage}{.56\linewidth}
\centering
\setlength\tabcolsep{1pt}
\begin{tabular}{c c>{\centering}p{1.5cm} ccc ccc}
\toprule Mode: &
\multicolumn{3}{c}{DAVIS First} &
\multicolumn{3}{c}{Dynamic Replica} &
% \multicolumn{3}{c}{100 support pts} &
% \multicolumn{3}{c}{1000 support pts}
\\
\cmidrule(lr){2-4}
\cmidrule(lr){5-7}
                    & AJ \ua        & \davg\ua      &  OA \ua        &
\davg\ua            & \davgvis\ua   & \davgocc\ua   \\
\midrule
no joint            & 55.6          & 70.1          &  83.0          &
54.4                & 62.4          & 28.8          \\
joint               & \textbf{62.2} & \textbf{75.7} &  \textbf{89.3}
                    & \textbf{61.6} & \textbf{68.9} &  \textbf{37.6} \\
\bottomrule
\end{tabular}
\caption{\textbf{Importance of joint tracking}.
We track a single target point with or without additional support points on DAVIS, and all target points separately or jointly on Dynamic Replica.
}%
\label{tab:tab_joint_tracking_ablation}
\end{minipage}%
\hspace{0.02\linewidth}
\begin{minipage}{0.40\linewidth}
\centering
\setlength\tabcolsep{1pt}
% \scriptsize

\begin{tabular}{cccc}
\toprule
\multirow{2}{*}{Unrolled Training} & \multicolumn{3}{c}{DAVIS First} \\
\cmidrule(lr){2-4}
                                   & AJ \ua                          &  \davg \ua      &  OA \ua        \\
\midrule
% \midrule
\xmark                             & 44.6                            &  60.5           &  75.3          \\
% 46.9                             & 69.1                            &  80.1           \\
\cmark                             & \textbf{62.2}                   &  \textbf{75.7}  &  \textbf{89.3} \\
% \textbf{60.6}                    & \textbf{75.4}                   &  \textbf{89.3 } \\
\bottomrule
\end{tabular}
\caption{\textbf{Unrolled training.}
\method is built for sliding window predictions.
Using them during training is important.
}%
\label{tab:tab_unrolling}
\end{minipage}
\end{table}

% \begin{tabular}{c c c>{\centering}p{1.5cm} ccc ccc}
% % ccc
% \toprule & Mode: &
% \multicolumn{3}{c}{DAVIS First} &
% \multicolumn{3}{c}{Dynamic Replica} &
% % \multicolumn{3}{c}{100 support pts} &
% % \multicolumn{3}{c}{1000 support pts}
% \\
% \cmidrule(lr){3-5}
% \cmidrule(lr){6-8}
%              &                    & AJ \ua        &  \davg\ua      &  OA \ua       &
%     \davg\ua & \davgvis\ua        & \davgocc\ua   \\
% \midrule
% (a)          & single             & 55.6          &  70.1          &  83.0         &
% 54.4         & 62.4               & 28.8          \\
% %            & 58.3               & 71.5          &  86.4          &  58.3         & 71.5 & 86.4& 58.3 & 71.5 & 86.4& 58.3 & 71.5 & 86.4
% (b)          & joint              & \textbf{62.2} &  \textbf{75.7} &  \textbf{89.3}
%              & \textbf{61.6}      & \textbf{68.9} &  \textbf{37.6} \\
% %            & 56.8               & 71.2          &  85.8          & 
% \bottomrule
% \end{tabular}
% \caption{\nk{combine with unrolled training?} \textbf{Importance of joint tracking}.
% We compare models pre-trained with time and cross-track attention, while tracking single or multiple target points, and using additional support points.
% }%
% \label{tab:tab_joint_tracking_ablation}

%% file: tabs/tab_ablate_clusters.tex
%\setlength\tabcolsep{.4em}
% \begin{table}[t]
% \centering
\begin{tabular}{cccccc}
\toprule
\multirow{2}{1cm}{\centering {\scriptsize\begin{tabular}[c]{@{}c@{}}Num.\\ proxy\\ tokens\end{tabular}}} &
\multicolumn{3}{c}{DAVIS First} &
\multirow{2}{1cm}{\centering {\scriptsize\begin{tabular}[c]{@{}c@{}}Max.\\ num.\\ tracks\end{tabular}}} &
\multirow{2}{*}{Time [s]}\\
\cmidrule(lr){2-4}
& AJ \ua& \davg\ua & OA\ua \\ 
\midrule
% \midrule
0  & 61.6 & 75.6 & 88.3 & $9.4$k &  207.3 \\
32 & 60.2 &  74.4 &  88.5 & $69.2$k & 26.8 \\
64 &  \textbf{62.2} & \textbf{75.7} & \textbf{89.3} & $69.2$k  & 27.9 \\
128 & 60.9 &  74.8 & 88.4 & $69.2$k & 30.1 \\
\bottomrule
\end{tabular}
\caption{The \textbf{proxy tokens} allow \method to scale. We train models with different numbers of proxy tokens and show that they help to decrease both memory complexity and inference time. 
We report the maximum number of tracks that can fit on a 80 GB GPU.}%
\label{tab:tab_ablate_clusters}
% \end{table}

%% file: tabs/tab_space_time_ablation.tex
% \begin{table}[t]
% \centering
\setlength{\tabcolsep}{1pt}

\begin{tabular}{c c c>{\centering}p{1.5cm} ccc }
% ccc
\toprule \multicolumn{2}{c}{Support:} &
\multicolumn{3}{c}{DAVIS First} &

\\
\cmidrule(lr){3-5}
% \cmidrule(lr){6-8}
global    &  local              & AJ \ua & \davg\ua & OA \ua  \\
    % \davg\ua & \davgvis\ua   & \davgocc\ua  \\
\midrule
\xmark & \xmark &    55.6  & 70.1 &     83.0   \\
\cmark & \xmark                & 56.8                  &  71.2              & 85.8          \\

\xmark  & \cmark              & 60.4 & 75.4 & 87.3 \\
\cmark  & \cmark & \textbf{62.2}         &  \textbf{75.7}     & \textbf{89.3} \\

\bottomrule
\end{tabular}
\caption{\textbf{Support points for TAP-Vid benchmarks}.
We track single target points from DAVIS using additional support points for context, as described in \cref{sec:point_selection}. Adding points for context always helps, while additional global \textit{and} local context works best.
}%
\label{tab:tab_joint_tracking_ablation_global_local}
%
% \end{table}

%% file: figure_lab.tex
\begin{figure*}[t]
\centering
\includegraphics[width=\linewidth]{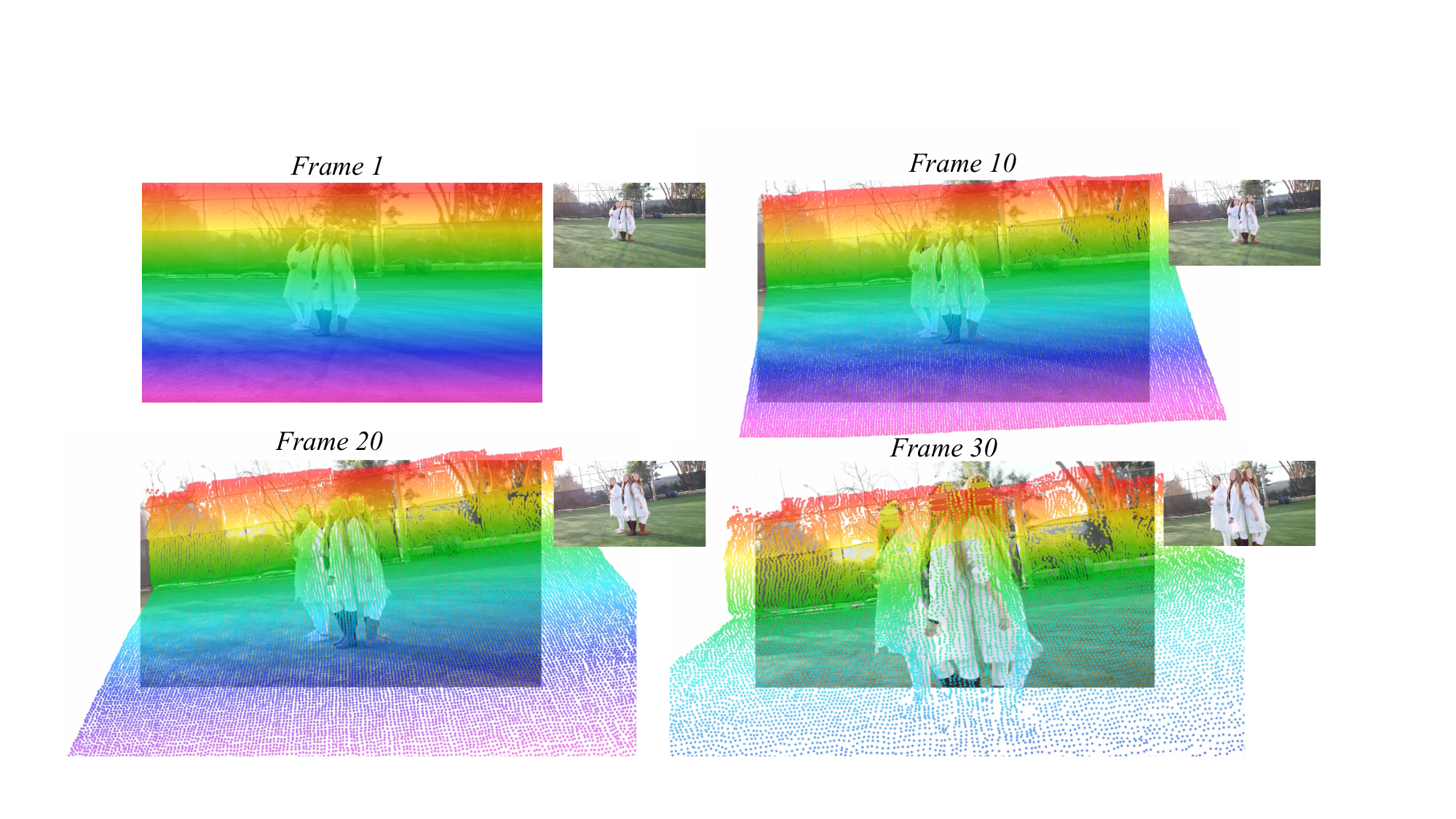}
% \fbox{\rule[-.5cm]{0cm}{4cm} \rule[-.5cm]{4cm}{0cm}}
\caption{\textbf{Qualitative Results}.
  \method can track points outside of the camera view. We visualize point displacements on a dense regular grid using an example from DAVIS~\cite{pont20172017}. In this example, the camera moves towards a group of people, while points in the background stay behind the camera view.}
\end{figure*}

%% file: new/5_limitations.tex
\subsection{Limitations}%
\label{s:limitations}

Despite its high performance, \method still occasionally makes tracking mistakes that a human would not make.
Since \method is trained on purely synthetic data, it sometimes does not generalize well to complex visual scenes with reflections and shadows. For example, \method tends to track shadows together with the objects that cast them.
Depending on the application this may be a desirable property (\eg video editing) or it may not be (\eg motion analysis).

%% file: new/6_conclusion.tex
\section{Conclusions}%
\label{s:conclusion}

We have presented \method, a transformer-based point tracker that tracks several points jointly, accounting for their dependencies.
\method is state-of-the-art on the standard tracking benchmarks, often by a substantial margin, can track through occlusions and when points leave the field of view, even for hundreds of frames, and can track a very large number of points simultaneously.
The transformer architecture is flexible and memory efficient. This allows for the integration of more functionalities in the future, such as 3D reconstruction.

%% file: new/7_acknowledgments.tex
\paragraph*{Acknowledgments}

We want to thank Laurynas Karazija for evaluating model efficiency, Luke Melas-Kyriazi and Jianyuan Wang for their paper comments, Roman Shapovalov, Iurii Makarov, Shalini Maiti, and Adam W.~Harley for the insightful discussions.
Christian Rupprecht was supported by ERC-CoG UNION101001212 and VisualAI EP/T028572/1.

%% file: supp_content.tex
\renewcommand{\thesubsection}{\Alph{subsection}}

% \appendix

\section*{Appendix}
\subsection{Additional ablations}
Here we provide additional ablation experiments that supplement the model choices from the main paper.

\paragraph{Support points.}

In addition to global and local support grids described in the paper, we also test using the SIFT detector~\cite{lowe04distinctive} to select support points in \cref{fig:accuracy_global_local_sift}.
While adding SIFT support points is better than sampling points uniformly on a global regular grid, adding a local grid yields significantly better results.
The combination of global and local grids is slightly better, and in  \cref{fig:accuracy_global_local} we further test this combination with different grid sizes ranging from 4 $\times$ 4 to 10 $\times$ 10.
We find that a configuration of 25 global and 64 local points is optimal for TAP-Vid-DAVIS in terms of speed and accuracy.
While there are differences in performance across these hyper-parameters, the overall impact on performance is less than one percentage point for a wide variety of configurations.
We use the selected configuration for evaluation on all the TAP-Vid benchmarks.

\begin{figure}[tb]
    \begin{minipage}{.47\linewidth}
        \centering
        \setlength\tabcolsep{0.9pt}
       \input{plot_global_local_sift}
    \end{minipage}%
    \hspace{0.02\linewidth}
    \begin{minipage}{0.5\linewidth}
        \centering
        \setlength\tabcolsep{1pt}
        \input{plot_global_local}
    \end{minipage}
\end{figure}

\paragraph{Training sliding window size.}

In, \cref{tab:tab_train_sliding_window_len_ablation} we ablate the sliding window size during training.
While \method can benefit from larger window sizes, the length of our training sequences from Kubric~\cite{tapvid} is limited to 24 frames.
Therefore, there is a trade-off between the context length and the ability to propagate predictions to future sliding windows.
A sliding window of 8 is optimal for our training dataset.

\paragraph{Inference sliding window size.}

In \cref{tab:tab_inf_sliding_window_len_ablation}, we examine the impact of the sliding window size on the performance of a model trained with a sliding window of $T=8$.
We find that the model performs best when the training and evaluation window sizes are identical.

\begin{table}[tb]
    \begin{minipage}{.47\linewidth}
        \centering
        \setlength\tabcolsep{0.9pt}
       \input{tabs/tab_train_sliding_window_size}
    \end{minipage}%
    \hspace{0.02\linewidth}
    \begin{minipage}{0.5\linewidth}
        \centering
        \setlength\tabcolsep{1pt}
        \input{tabs/tab_inf_sliding_window_len_ablation}
    \end{minipage}
\end{table}

\subsection{Efficiency}

In \cref{fig:efficiency} we evaluate the efficiency of PIPs++~\cite{zheng2023pointodyssey}, TAPIR~\cite{doersch2023tapir} and \method on a 50-frame video of resolution $256\times256$ by running them on an A40 GPU\@.
We track points sampled on a regular grid at the first frame of the video and report the time it takes to process the entire video sequence.
CoTracker is slower than TAPIR but faster than PIPs++ (while achieving better accuracy than both).

\subsection{Evaluation on TAP-Vid-Kinetics}

In \cref{tab_kinetics} we evaluate \method on TAP-Vid-Kinetics, a dataset of 1144 videos of approximately 250 frames each from Kinetics~\cite{carreira2017quo}.
Some of these videos are discontinuous, i.e., they are composed of continuous video chunks.
\method is designed for continuous videos, and TAPIR contains a matching stage inspired by TAP-Net~\cite{tapvid}, which can help with such combined video chunks.
This is why the performance gap between TAPIR and \method for this benchmark is smaller than for other benchmarks.

\begin{table}[tb]
    \begin{minipage}{.47\linewidth}
        \centering
        \setlength\tabcolsep{0.9pt}
        \input{tabs/tab_kinetics}

    \end{minipage}%
    \hspace{0.02\linewidth}
    \begin{minipage}{0.5\linewidth}
        \centering
        \setlength\tabcolsep{1pt}
        \input{tabs/tab_model_stride_ablation}
    \end{minipage}
\end{table}
\begin{figure}[tb]
    \begin{minipage}{.47\linewidth}
        \centering
        \setlength\tabcolsep{0.9pt}
       \input{plot_speed}
    \end{minipage}%
    \hspace{0.02\linewidth}
    \begin{minipage}{0.5\linewidth}
        \centering
        \setlength\tabcolsep{1pt}
        \input{plot_ablate_updates}
    \end{minipage}
\end{figure}
\subsection{Implementation Details}

In this section, we complete the description of implementation details from the main paper.
We will release the code and models with the paper.

\paragraph{Feature CNN.}

Given a sequence of $T'$ frames with a resolution of 384$\times$512, we compute features for each frame using a 2-dimensional CNN\@.
This CNN downsamples the input image by a factor of 4 (feature stride) and outputs features with 128 channels.
Our CNN is the same as the one used in PIPs~\cite{pips}.
It consists of one $7\times 7$ convolution with a stride of 2, eight residual blocks with $3\times 3$ kernels and instance normalization, and two final convolutions with $3\times 3$ and $1 \times 1$ kernels.
In \cref{tab:tab_model_stride_ablation}, we train \method with two different feature downsampling factors (strides).

\paragraph{Sliding windows.}

When passing information from one sliding window to the next, we concatenate binary masks of shape $(N, T)$ with visibility logits that indicate where the model needs to make predictions.
For example, masks in the first sliding window would be equal to 1 from the frame where we start tracking the point, and 0 before that.
Masks for all the subsequent sliding windows will be equal to 1 for the first overlapping $T/2$ frames, and 0 for the remaining $T/2$.
During training, tracking starts either from the first or from a random frame where the point is visible.
If a point is not yet visible in the current sliding window, it will be masked out during the application of cross-track attention.

\paragraph{Iterative updates.}

We train the model with $M=4$ iterative updates and evaluate it with $M=6$.
This setting provides a reasonable trade-off between speed and accuracy, as evaluated in \cref{fig:plot_ablate_updates}. Interestingly, the performance remains stable for 4--8 iterative updates and begins to degrade slowly thereafter.

\paragraph{Training.}

\method is trained with a batch size of 32, distributed across 32 GPUs.
After applying data augmentations, we randomly sample 768 trajectories for each batch, with points visible either in the first or in the middle frame.
We train the model for 50,000 iterations with a learning rate of $5e^{-4}$ and a linear 1-cycle~\cite{smith2019super} learning rate schedule, using the AdamW~\cite{loshchilov2017decoupled} optimizer.

\paragraph{Augmentations.}

During training, we employ Color Jitter and Gaussian Blur to introduce color and blur variations.
We augment occlusions by either coloring a randomly chosen rectangular patch with its mean color or replacing it with another patch from the same image.
Random scaling across height and width is applied to each frame to add diversity.

\subsection{Broader societal impact}

Motion estimation, whether in the guise of point tracking or optical flow, is a fundamental, low-level computer vision task.
Many other tasks in computer vision build on it, from 3D reconstruction to video object segmentation.
Ultimately, motion estimation algorithms are important components of a very large number of applications of computer vision in many different areas.
Point tracking has no \emph{direct} societal impact; however, positive or negative effects on society can materialise through its use in other algorithms, depending on the final application.

\subsection{Technical note}

We work out the coordinate mappings between layers in a neural network.
We assume that layers are 1D, as the generalization to several dimensions is immediate.

Consider a tensor $x \in \mathbb{R}^{W}$ with $W_1$ components indexed by $i\in \Omega = \{0,\dots,W_1-1\}$.
We also interpret each element in the tensor as a unit tile, collectively covering the interval $U = [0,W]$, and we denote via $u \in U$ the corresponding continuous coordinate.
Mapping each index to the center of the corresponding tile, we have the correspondence:
$$
\Omega \rightarrow U,
~~~
i \mapsto u(i) = i + \frac{1}{2}.
$$

We interpret the tensor $x$ as providing information about image points $u \in U$.
By reading off element $x_i$, we obtain information about location $u(i)$.

\paragraph{Layers and transformations.}

Next consider two tensors $x_1 = \phi(x_2)$ related by a neural network layer $\phi$.
The layer establishes a mapping $i_1(i_2)$, or equivalently $u_1(u_2)$, between indices/coordinates of the two tensors.
We write $i_1 = \alpha_{12} i_2 + \beta_{12}$, so that the following commutative diagram holds:
\begin{center}
\begin{tikzcd}
u_2 \in U_2
\arrow{d}[swap]{\alpha_{12} u_2 + \beta_{12} + \frac{1 - \alpha_{12}}{2}}  &
i_2 \in \Omega_2
\arrow{d}{\alpha_{12} i_2 + \beta_{12}}
\arrow{l}{+\frac{1}{2}} \\
u_1 \in U_1 & i_1 \in \Omega_1
\arrow{l}[swap]{+\frac{1}{2}}
\end{tikzcd}
,
\end{center}
If we have chains of layers $x_0 \mapsto x_1 \mapsto x_2$, we can chain the corresponding diagrams:
\begin{center}
\begin{tikzcd}
u_2 \in U_2
\arrow{d}[swap]{\alpha_{12} u_2 + \beta_{12} + \frac{1 - \alpha_{12}}{2}}
&
i_2 \in \Omega_2
\arrow{d}{\alpha_{12} i_2 + \beta_{12}}
\arrow{l}{+\frac{1}{2}}
\\
u_1 \in U_1
\arrow{d}[swap]{\alpha_{1} u_1 + \beta_{1} + \frac{1 - \alpha_{1}}{2}}
& i_1 \in \Omega_1
\arrow{l}[swap]{+\frac{1}{2}}
\arrow{d}{\alpha_{1} i_1 + \beta_{1}}
\\
u_0 \in U_0
& i_0 \in \Omega_0
\arrow{l}[swap]{+\frac{1}{2}}
\end{tikzcd}
,
\end{center}
where for simplicity we denote $\alpha_{0i} = \alpha_i$ and $\beta_{0i} = \beta_i$.
With this, we find the recurrent equation:
$$
\boxed{
\alpha_n = \alpha_{n-1} \alpha_{n-1,n},
~~~
\beta_n = \beta_{n-1} + \alpha_{n-1} \beta_{n-1,n}
}
$$
so that we can always refer any layer back to the input image as follows:
$$
i_0 = \alpha_n i_n + \beta_n,~~~
u_0 = \alpha_n u_n + \beta_n + \frac{1-\alpha_{n}}{2}.
$$

Finally, we consider a sampling layer, implemented using, e.g., bilinear interpolation.
Applied to an input tensor $x_1$, we can write the latter as reading off values $x_2(u_2)$ at designated coordinates $u_2$.
This is similar to computing an output tensor $x_2$ as before, but with the ability of reading off values at arbitrary coordinates $u_2$ instead of fixed indices $i_2$.
This is written down as a function
$
x_2(u_2) =\phi(x_1, u_2).
$
Just like before, we write correspondences:
$$
i_1 = \alpha_{12} i_2 + \beta_{12},~~~
u_1 = \alpha_{12} u_2 + \beta_{12} + \frac{1-\alpha_{12}}{2}
$$
although, obviously, the index $i_2 = u_2 - 1/2$ is ``virtual''.

\paragraph{Filter-like layers.}

Next consider two tensors $x_1 = \phi(x_2)$ where $\phi$ is a neural network layer.
Assume that the layer is a filter with filter size $F$, padding $P$ and stride $S$ (almost all layers in a neural network are of this kind, often special cases of this).
Pixel $i_2$ is obtained by combinations of the values of pixels $i_1 \in \{ i_2 S - P,\dots,i_2 S - P + F-1 \}$ of the input.
Hence, we can assume that the pixel of index $i_2$ corresponds to the center of this range, which has coordinate:
$$
u_1(i_1)
= \frac{
\left(u_1(i_2 S - P) - \frac{1}{2}\right) +
\left(u_1(i_2 S - P + F-1) + \frac{1}{2}\right)
}{2}
= i_2 S - P + \frac{F-1}{2} + \frac{1}{2}.
$$
Hence:
$$
\boxed{\alpha_{12} = S,~~~
\beta_{12} = \frac{F-1}{2} - P}.
$$
Note that we can make $\beta_{12} = 0$ by choosing $P = (F-1)/2$ (which requires odd-sized filters).
By using the recurrence equation above, with a chain of such layers we obtain:
$$
\alpha_n = \prod_{i=1}^n S_n,~~~
\beta_n = 0.
$$

\paragraph{Interpolation layers.}

Next, we consider the interpolation layer $x_2 = \phi(x_1)$.
With the \texttt{align\_corners} option set to \texttt{True}, pixels $i_1=0,W_1-1$ are mapped to pixels $i_2=0,W_2-1$.
Hence:
$$
i_1
= i_2 \frac{W_1-1}{W_2-1}
= \alpha_{12}i_2 + \beta_{12}.
$$
Hence we have:
$$
\alpha_{12} = \frac{W_1-1}{W_2-1},~~~
\beta_{12} = 0.
$$
With the \texttt{align\_corners} option set to \texttt{False}, coordinates $u_1=0,W_1$ are mapped to coordinates $u_2=0,W_2$.
Hence:
$$
u_2
= u_1 \frac{W_1}{W_2}
= \alpha_{12} u_1
+ \beta_{12}
+ \frac{1 - \alpha_{12}}{2}.
$$
Hence:
$$
\alpha_{12} = \frac{W_1}{W_2},~~~
\beta_{12} = - \frac{1 - \alpha_{12}}{2}
= \frac{W_1-W_2}{2W_2}.
$$

\paragraph{Sampling layers.}

Next, we figure out the sampling layer $x_2(u_2) = \phi(x_1,u_2)$.
When we use bilinear sampling to sample features from $x_1$, we use the \texttt{grid\_sample} operator with \texttt{align\_corners} set to either \texttt{True} or \texttt{False}.

We can model this as introducing a coordinate $u_2 \in U_2 = [-1,1]$ with the following mappings.
If \texttt{align\_corners} is \texttt{False}, we align the extrema $-1,1$ of the output range to the edges of the input image, so that:
$$
u_1(u_2)
=
\frac{W_1}{2} (u_2 + 1)
=
\alpha_{12} u_1
+ \beta_{12}
+ \frac{1 - \alpha_{12}}{2}.
$$
i.e.,
$$
\alpha_{12} = \frac{W_1}{2},~~~
\beta_{12} = \frac{W_1}{2} - \frac{1 - \alpha_{12}}{2} = - \frac{1}{2} - \frac{W_1}{4}.
$$
If \texttt{align\_corners} is \texttt{True}, we align the extrema of the output range $-1, 1$ with the centers of the input edge pixels, so that:
$$
i_1(u_2)
=
\frac{W_1-1}{2} (u_2 + 1)
=
\alpha_{12}u_2 + \beta_{12},
$$
i.e.,
$$
\alpha_{12} = \frac{W_1 - 1}{2},~~~
\beta_{12} = \frac{W_1 - 1}{2}.
$$

\paragraph{Sampling layer, alt.~conventions.}

In practice, we redefine the mapping layers in such a way that, if \texttt{align\_corners} is \texttt{False}, then $u_2 \in U_2 = [0, W_1]$ and the extrema $0, W_1$ align edges of the input image, so that:
$$
u_1(u_2)
=
u_2
=
\alpha_{12} u_1
+ \beta_{12}
+ \frac{1 - \alpha_{12}}{2}.
$$
i.e.,
$$
\alpha_{12} = 1,~~~
\beta_{12} = 0,
$$
which means:
$$
u_1(u_2) = u_2, ~~~
i_1(i_2) = i_2.
$$
With \texttt{align\_corners} is \texttt{True}, then $u_2 \in U_2 = [0, W_1-1]$ and we align the extrema $0, W_1-1$ to the center of the edge pixels, so that:
$$
i_1(u_2)
=
u_2
=
\alpha_{12}u_2 + \beta_{12} - \frac{1-\alpha_{12}}{2} - \frac{1}{2},
$$
i.e.,
$$
\alpha_{12} = 1,~~~
\beta_{12} = \frac{1}{2},
$$
which means:
$$
u_1(u_2) = u_2 + \frac{1}{2},~~~
i_1(i_2) = i_1 + \frac{1}{2}.
$$

%% file: plot_global_local_sift.tex
\includegraphics[width=\textwidth]{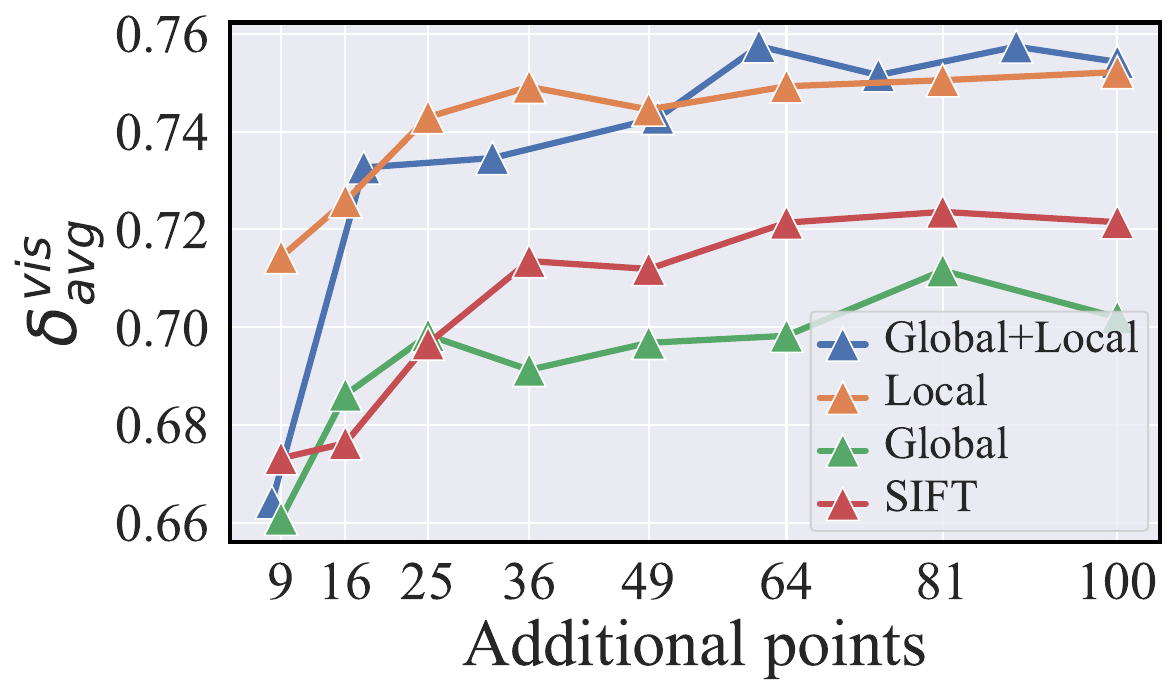}
\caption{\textbf{Accuracy on TAP-Vid-DAVIS} depending on the number of additional local, global grid and SIFT support points. When combining point sources, we use (close to) the same number of points.}%
\label{fig:accuracy_global_local_sift}

%% file: plot_global_local.tex
\includegraphics[width=0.9\textwidth]{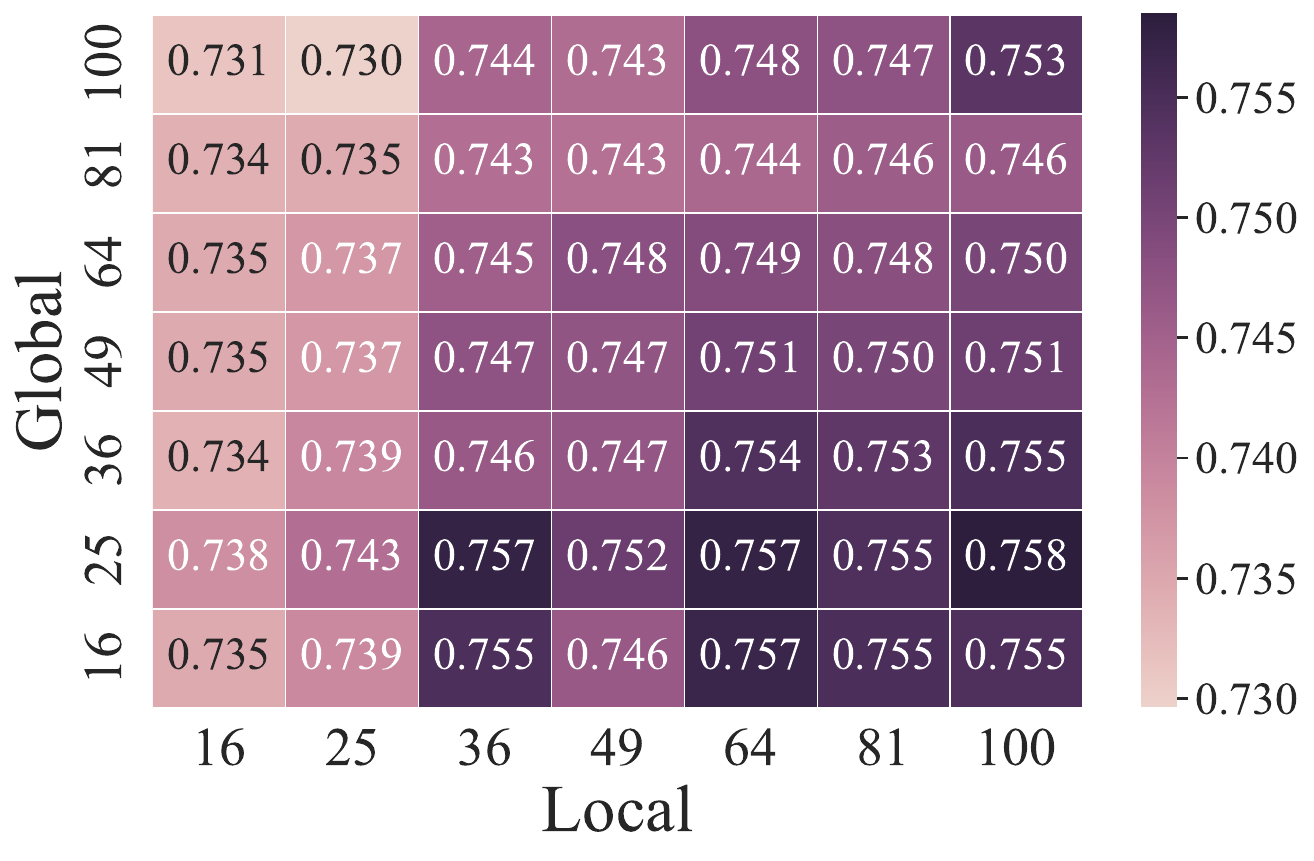}
\caption{\textbf{Number of support points.} Accuracy for additional global grid and local grid points. We compute $\delta_{avg}^{vis}$ on TAP-Vid-DAVIS to find the optimal grid sizes for TAP-Vid benchmarks. We use 25 \& 64 because of speed and higher occlusion accuracy ($89.3$ vs. $88.7$).}%
\label{fig:accuracy_global_local}

%% file: tabs/tab_train_sliding_window_size.tex
%\setlength\tabcolsep{.4em}
% \begin{table}[t]
% \centering
% \footnotesize

\begin{tabular}{ ccccc }
\toprule
 \multirow{2}{*}{Window size}  & \multicolumn{3}{c}{DAVIS First} \\
\cmidrule(lr){2-4}
% \cmidrule(lr){5-6}
            & AJ            & \! \! \! $< \delta^x_{avg}$\! & OA    \\
\midrule

4         &  56.7  & 73.1 &  86.5  \\
\textbf{8}         &  \textbf{62.2}          & \textbf{75.7}  & \textbf{89.3} \\
16 &  61.1 & 75.5 & 88.4 \\
\bottomrule
\end{tabular}
\caption{  \textbf{Training sliding window size}. We train and evaluate \method with varying window lengths. As the training data only contains sequences of length 24, the model does not benefit from training with a bigger sliding window.}\label{tab:tab_train_sliding_window_len_ablation}
% \end{table}

%% file: tabs/tab_inf_sliding_window_len_ablation.tex
\begin{tabular}{ cccc }
\toprule
 \multirow{2}{*}{Window size} & \multicolumn{3}{c}{DAVIS First} \\
\cmidrule(lr){2-4}
                              & AJ                              &  \! \! \! $< \delta^x_{avg}$\! & OA            \\
\midrule
4                             & 54.9                            &  70.8                          & 83.2          \\
\textbf{8}                    & \textbf{62.2}                   &  \textbf{75.7}                 & \textbf{89.3} \\
16                            & 54.6                            &  68.8                          & 82.9          \\
\bottomrule
\end{tabular}
\caption{\textbf{Inference sliding window size}. The model is trained with sliding window size $T=8$.
The same window size gives the best results.}%
\label{tab:tab_inf_sliding_window_len_ablation}
\vspace{2em}

%% file: tabs/tab_kinetics.tex
\begin{tabular}{cccc}
\toprule
 \multirow{2}{*}{Stride} &  \multicolumn{3}{c}{Kinetics First} \\
\cmidrule(lr){2-4}
                         &  AJ                                 &  $\delta^{vis}_{avg}$ &  OA           \\
\midrule
PIPs++                   &  ---                                &  58.5                 &  ---          \\
TAPIR                    &  \textbf{49.6}                      &  64.2                 &  85.0         \\
CoTracker (Ours)         &  48.8                               &  \textbf{64.5}        &  \textbf{85.8}
                         \\ \bottomrule
\end{tabular}
\caption{
\textbf{Evaluation on TAP-Vid-Kinetics}. Kinetics contains videos with multiple shots, which does not satisfy the assumptions of most trackers, including CoTracker. The results of CoTracker and TAPIR are comparable, even though TAPIR is an offline method with a specifically designed matching module.
}\label{tab_kinetics}
\vspace{2em}

%% file: tabs/tab_model_stride_ablation.tex
\begin{tabular}{cccc}
\toprule
\multirow{2}{*}{Stride} & \multicolumn{3}{c}{DAVIS First} \\
\cmidrule(lr){2-4}
                        & AJ            & $\delta^{vis}_{avg}$ &  OA            \\
\midrule
 8                      & 52.5          & 68.6                 &  85.7          \\

\textbf{4}              & \textbf{62.2} & \textbf{75.7}        &  \textbf{89.3} \\
\bottomrule
\end{tabular}
\caption{
\textbf{Feature stride ablation}. We ablate the stride of the Feature CNN\@.
Higher resolution features help to make much more accurate predictions.}%
\label{tab:tab_model_stride_ablation}
\vspace{5.5em}

%% file: plot_speed.tex
\includegraphics[width=\textwidth]{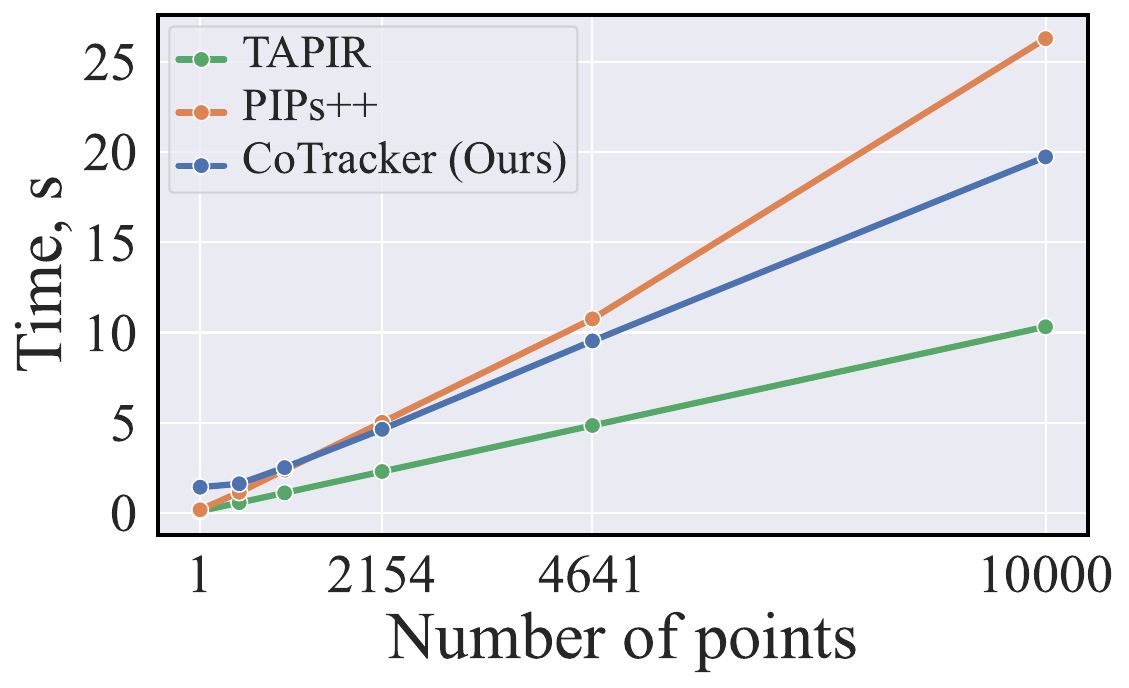}
\caption{\textbf{Efficiency}. We track N points sampled on the first frame of a 50-frame video of resolution $256 \times 256$ using a NVIDIA A40 48GB GPU, and report the time it takes the method to process the video.
}%
\label{fig:efficiency}

%% file: plot_ablate_updates.tex
\includegraphics[width=\textwidth]{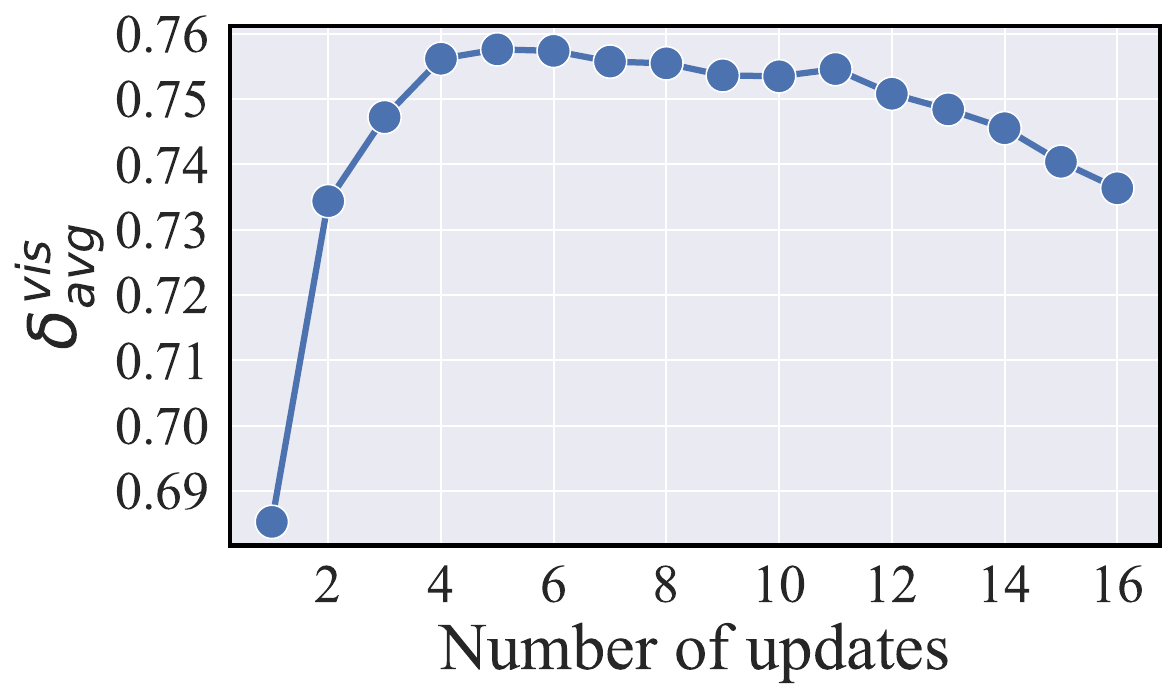}
\caption{\textbf{Inference iterative updates.} We ablate the number of iterative updates $M$ during inference. The network is trained with $M=4$.}%
\label{fig:plot_ablate_updates}
\vspace{2.5em}